\documentclass[a4paper,11pt]{article}

\usepackage{graphicx}  
\usepackage{url}       
\usepackage{times}
\usepackage{natbib}
\usepackage[margin=25mm]{geometry}

\usepackage[colorinlistoftodos, textsize=tiny]{todonotes}
\usepackage{amsmath}
\usepackage{amssymb}
\usepackage{amsfonts}
\usepackage{bm}
\usepackage{diagbox}
\usepackage{graphicx, color}
\usepackage{algorithm}
\usepackage{multirow}
\usepackage{subcaption}
\usepackage[noend]{algpseudocode}

\date{}

\title{The Fast and the Flexible: training neural networks to learn to follow instructions from 
small data}
\author{Rezka Leonandya\\
 University of Amsterdam\\
 rezka.aufar@gmail.com\\
 \and Dieuwke Hupkes\\
 University of Amsterdam\\
 d.hupkes@uva.nl\\
 \and Elia Bruni\\
 University of Amsterdam\\
 elia.bruni@gmail.com\\
 \and Germ\'an Kruszewski\\
 Facebook AI\\
 germank@gmail.com
 }

\makeatletter
\newcommand\nocaption{%
    \renewcommand\p@subfigure{}
    \renewcommand\thesubfigure{\thefigure\alph{subfigure}}
}
\makeatother

\date{}

\begin{document}
\maketitle
\thispagestyle{empty}
\pagestyle{empty}
\newcommand{\newcite}[1]{\citet{#1}}

\begin{abstract}

    Learning to follow human instructions is a long-pursued goal in artificial intelligence. 
    The task becomes particularly challenging if no prior knowledge of the employed language is assumed while relying only on a handful of examples to learn from.
    Work in the past has relied on hand-coded components or manually engineered features to provide strong inductive biases that make learning in such situations possible.
    In contrast, here we seek to establish whether this knowledge can be acquired automatically by a neural network system through a two phase training procedure: A (slow) offline learning stage where the network learns about the general structure of the task and a (fast) online adaptation phase where the network learns the language of a new given speaker.
    Controlled experiments show that when the network is exposed to familiar instructions but containing novel words, the model adapts very efficiently to the new vocabulary. Moreover, even for human speakers whose language usage can depart significantly from our artificial training language, our network can still make use of its automatically acquired inductive bias to learn to follow instructions more effectively.
\end{abstract}

\section{Introduction}

\begin{figure}[tb]
    \centering
    \includegraphics[width=170pt]{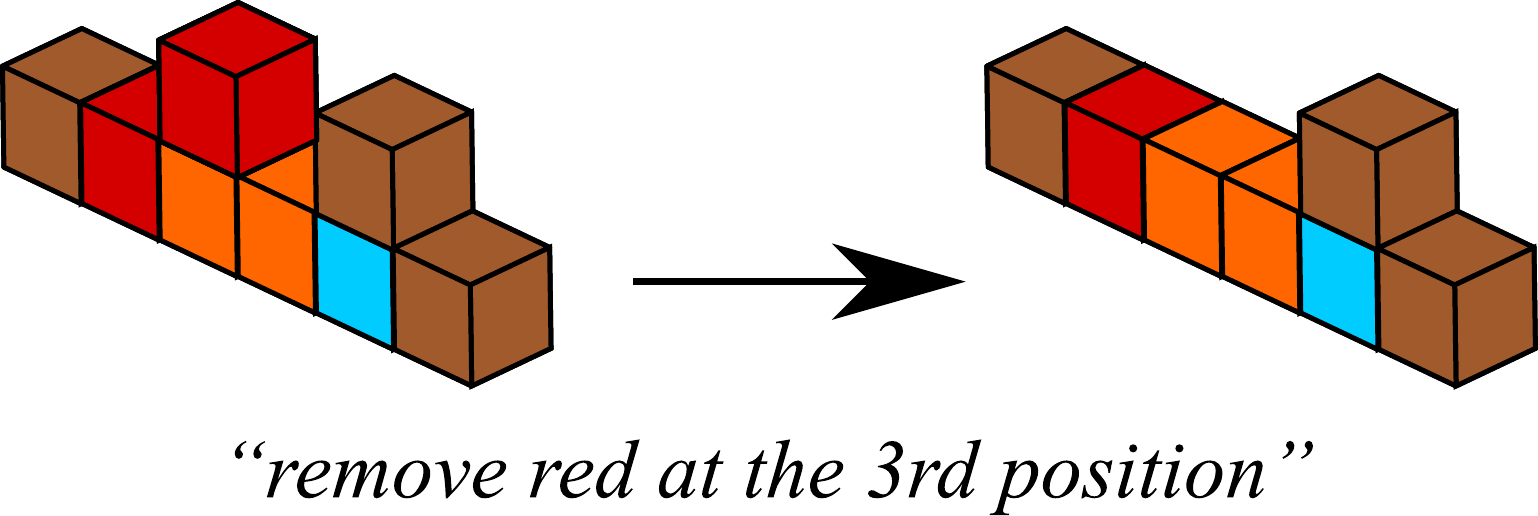}
    \caption{Illustration of the SHRDLURN task of \cite{Wang:etal:2016}\label{fig:task-cartoon}}
\end{figure}

 Learning to follow instructions from human speakers is a long-pursued goal in artificial intelligence, tracing back at least to Terry Winograd's work on SHRDLU \citep{Winograd:1972}. 
 This system was capable of interpreting and following natural language instructions about a world composed of geometric figures.
 While this first system relied on a set of hand-coded rules to process natural language, most of recent work aimed at using machine learning to map linguistic utterances into their semantic interpretations \citep{Chen:Mooney:2011,Artzi:Zettlemoyer:2013,Andreas:Klein:2015}. 
 Predominantly, they assumed that users speak all in the same natural language, and thus the systems could be trained offline once and for all.
 However, recently \cite{Wang:etal:2016} departed from this assumption by proposing SHRDLURN, a coloured-blocks manipulation language game.
 There, users could issue instructions in any arbitrary language to a system that must incrementally learn to interpret it (see Figure \ref{fig:task-cartoon} for an example). 
 This learning problem is particularly challenging because human users typically provide only a handful of examples for the system to learn from.
 Therefore, the learning algorithms must incorporate strong inductive biases in order to learn effectively.
 That is, they need to complement the scarce input with priors that would help the model make the right inferences even in the absence of positive data.
 A way of giving the models a powerful inductive bias is by hand-coding features or operations that are specific to the given domain where the instructions must be interpreted. 
For example, \cite{Wang:etal:2016} propose a log-linear semantic parser which crucially relies on a set of hand-coded functional primitives. 
 While effective, this strategy severely curtails the portability of a system: For every new domain, human technical expertise is required to adapt the system.
 Instead, we would like these inductive biases to be learned automatically without human intervention.
 That is, humans should be free from the burden of thinking what are useful 
 primitives for a given domain, but still obtain systems that can learn fast
 from little data.

In this paper, we introduce a neural network system that learns domain-specific priors directly from data. This system can then be used to quickly learn the language of new users online. It uses a two phase regime: First, the network is trained \textit{offline} on easy-to-produce artificial data to learn the mechanics of a given task.
Next, the network is deployed to real human users who will train it \textit{online} with just a handful of examples. 
While this implies that some of the manual effort needed to design useful
primitive functions would go in developing the artificial data, we envision 
that in many real-world situations it could be easier to provide examples of
expected interactions than thinking of what could be useful primitives 
involved in them.
On controlled experiments we show that our system can recover the meaning of sentences where some words where scrambled, even though it does not display evidence of compositional learning. On the other hand, we show that the offline training phase allows it to learn faster from limited data, compared to a neural network system that did not go through this pre-training phase. 
We hypothesize that this system learns useful inductive biases, such as the types of operations that are likely to be requested.
In this direction, we show that the performance of our best-performing system correlates with that of \citeauthor{Wang:etal:2016}, where these operations were encoded by hand.

The work in this paper is organized as follows: We first start by creating a large artificially generated dataset to train the systems in the offline phase. We then experiment with different neural network architectures 
to find which general learning system adapts best for this task%
. Then, we propose how to adapt this network by training it online %
and confirm its effectiveness on recovering the meaning of scrambled words %
and on learning to process the language from human users, using the dataset introduced by \newcite{Wang:etal:2016}.

\section{Related Work}
Learning to follow human natural language instructions has a long tradition in
NLP, dating at least back to the work of Terry Winograd \cite{Winograd:1972}, who developed a rule-based
system for this endeavour. Subsequent work centered around automatically
learning the rules to process language
\citep{Shimizu:etal:2009,Chen:Mooney:2011,Artzi:Zettlemoyer:2013,Vogel:Jurafsky:2010,Andreas:Klein:2015}.
This line of work assumes that users speak all in the same language, and thus a
system can be trained on a set of dialogs pertaining to some of those speakers
and then generalize to new ones. 
Instead, \cite{Wang:etal:2016} describe a block manipulation game in which a
system needs to learn to follow natural language instructions produced by human
users using the correct outcome of the instruction as feedback. What
distinguishes this from other work is that every user can speak in their own
--natural or invented-- language. For the game to remain engaging, the system 
needs to quickly adapt to the user's language, thus requiring a system that can learn much
faster from small data.  The system they propose is composed of a set of
hand-coded primitives (e.g., \texttt{remove, red, with}) that can manipulate
the state of the block piles and a log-linear learning model that learns to map
n-gram features from the linguistic instructions (like, for instance,
`\textit{remove red}') to expressions in this programming language (e.g.,
\texttt{remove(with(red))}).
Our work departs from this base in that we provide no hand-coded primitives to
solve this task, but aim at learning an end-to-end system that follows natural
language instructions from human feedback. 
Another line of research that is closely related to ours, is that of \emph{fast 
mapping} \citep{Lake:etal:2011,Trueswell:etal:2013,Herbelot:Baroni:2017}, 
where the goal is to acquire a new concept from a single example of its usage 
in context. While we don't aim at learning new concepts here, we do want to 
learn from few examples to draw an analogy between a new term and a previously 
acquired concept.
Finally, our work can be seen as an instance of the transfer learning paradigm
\citep{Pan:etal:2010}, which has been successful in both linguistic \citep{Mikolov:etal:2013,Peters:etal:2018}
and visual processing \citep{Oquab:etal:2014}.  Rather than transferring knowledge from one task
to another, we are transferring between artificial and natural data.

\section{Method}
A model aimed at following natural language instructions must master at least two skills. First, it needs to process the language of the human user. Second, it must act on the target domain in sensible ways (and not trying actions that a human user would probably never ask for). 
Whereas the first aspect is dependent on each specific user's language, the second requirement is not related to a specific user, and could -- as illustrated by the successes of \citeauthor{Wang:etal:2016}'s log-linear model -- be learned beforehand.
To allow a system to acquire these skills automatically from data, we introduce a two-step training regime. 
First, we train the neural network model offline on a dataset that mimics the target task. 
Next, we allow this model to independently adapt to the language of each particular human user by training it online with the examples that each user provides.

\subsection{Offline learning phase}
\label{sec:offline-training}

The task at hand is, given a list of piles of coloured blocks and a natural language instruction, to produce a new list of piles that matches the request.
The first step of our method involves training a neural network model to perform this task.
We used supervised learning to train the system on a dataset that we constructed by simulating a user playing the game.
In this way, we did not require any real data to kick-start our model. 
Below we describe, first, the procedure used to generate the dataset and, second, the neural network models that were explored in this phase.

\paragraph{Data} 
The data for SHRDLURN task takes the form of triples: a start configuration of colored blocks grouped into piles, a natural language instruction given by a user and the resulting configuration of colored blocks that comply with the given instruction\footnote{The original paper produces a rank of candidate configurations to give to a human annotator. Since here we focus on pre-annotated data where only the expected target configuration is given, we will restrict our evaluation to top-1 accuracy.} (Figure \ref{fig:task-cartoon}).
We generated 88 natural language instructions following the grammar in Figure \ref{fig:grammar}. 
The language of the grammar was kept as minimal as possible, with just enough variation to capture the simplest possible actions in this game.
Furthermore, we sampled as many as needed initial block configurations by building 6 piles containing a maximum of 3 randomly sampled colored blocks each. 
The piles in the dataset were serialized into a sequence by encoding them into 6 lists delimited by a special symbol, each of them containing a sequence of color tokens or a special \texttt{\small empty} symbol.
We then computed the resulting target configuration using a rule-based interpretation of our grammar. 
An example of our generated data is depicted in Figure~\ref{fig:ds-example}.

\renewcommand{\ttdefault}{lmtt} 
\begin{figure}
    \begin{subfigure}[b]{0.5\textwidth}
    \texttt{
    \begin{footnotesize}
    \begin{center}
	\begin{tabular}{ccc}
		S & $\rightarrow$ & VERB COLOR at POS tile \\
		VERB & $\rightarrow$ & add $\vert$ remove \\
		COLOR & $\rightarrow$ & red $\vert$ cyan $\vert$ brown $\vert$ orange \\
		POS & $\rightarrow$ & 1st $\vert$ 2nd $\vert$ 3rd $\vert$ 4th $\vert$ \\
        & & 5th $\vert$ 6th $\vert$ even $\vert$ odd $\vert$ \\
        & & leftmost $\vert$ rightmost $\vert$ every \\
	\end{tabular}
    \end{center}
    \end{footnotesize}
    }
	\caption{Grammar of our artificially generated language\label{fig:grammar}}
    \end{subfigure}
    \begin{subfigure}[b]{0.4\textwidth}
\small
\begin{center}
\begin{tabular}{ll}
Instruction & \tt remove red at 3rd tile \\
Initial Config. & \scriptsize \tt BROWN X X \# RED X X \# ORANGE RED X \\
Target Config. & \scriptsize \tt BROWN X X \# RED X X \# ORANGE X X \\
\end{tabular}
\end{center}
\caption{Example of an entry in our dataset. We show three rather than six 
columns for conciseness.}\label{fig:ds-example}
    \end{subfigure}
    \caption{Artificially generated data}
\end{figure}

\paragraph{Model} To model this task we used an encoder-decoder \citep{Sutskever:etal:2014} architecture: The encoder reads the natural language utterance $\mathbf{w}=w_1, \dots, w_m$ and transforms it into a sequence of feature vectors $\mathbf{h_1}, \dots, \mathbf{h}_m$, which are then read by the decoder through an attention layer. 
This latter module reads the sequence describing the input block configurations $\mathbf{x}=x_1, \dots, x_n$ and produces a new sequence $\mathbf{\hat{y}}=\hat{y}_1, \dots, \hat{y}_n$ that is construed as the resulting block configuration.
To pass information from the encoder to the decoder, we equipped the decoder with an attention mechanism \citep{Bahdanau:etal:2014,Luong:etal:2015}. 
This allows the decoder, at every timestep, to extract a convex combination of the hidden vectors $\mathbf{h}$.
We trained the system parameters $\theta = (\theta_{\text{enc}}, \theta_{\text{dec}})$ so that the output matches the target block configuration $\mathbf{y}=y_1, \dots, y_n$ (represented as 1-hot vectors) using a cross-entropy loss:
\begin{align*}
    \mathbf{h} &= \operatorname{encoder}_{\theta_{\text{enc}}}(\mathbf{w})\\ 
    \mathbf{\hat{y}} &= \operatorname{decoder}_{\theta_{\text{dec}}}(\mathbf{x}|\mathbf{h}) \\
    \mathcal{L}(\mathbf{y},\mathbf{\hat{y}}) &= \sum_{i=1}^{n} y_i \log \hat{y}_i 
\end{align*}
Both the encoder and decoder modules are sequence models, meaning that they read a sequence of inputs and compute, in turn, a sequence of outputs, and that can be trained end-to-end.
We experimented with two state-of-the-art sequence models: A standard recurrent LSTM \citep{Hochreiter:Schmidhuber:1997} and a convolutional sequence model \citep{Gehring:etal:2016,Gehring:etal:2017}, which has been shown to outperform the former on a range of different tasks \citep{Bai:etal:2018}. 
For the convolutional model we used kernel size $k=3$ and padding to make the size of the output match the size of the input sequence. 
Because of the invariant structure of the block configuration that is organized into lists of columns, we expected the convolutional model (as a decoder) to be particularly well-fit to process them.
We explored all possible combinations of architectures for the encoder and
decoder components.  
Furthermore, as a simple baseline, we also considered a bag-of-words encoder that computes the average of trainable word embeddings.

\subsection{Online learning phase}
\label{sec:method-online}

Once the model has been trained to follow a specific set of instructions given by a simulated user, we want it to serve a new, real user, who does not know anything about how the model was trained and is encouraged to communicate with the system using her own language. 
To do so, the model will have to adapt to follow instructions given in a potentially very different language from the one it has seen during offline training. 
One of the first challenges it will encounter is to quickly master the meaning of new words.  
This challenge of inferring the meaning of a word from a single exposure goes by the name of `fast-mapping' \citep{Lake:etal:2011,Trueswell:etal:2013}. 
Here, we take inspiration from the method proposed by \cite{Herbelot:Baroni:2017}, who learn the embeddings for new words with gradient descent, freezing all the other network weights. 
We further develop it by experimenting with different variations of this method: Like them, we try learning only \textit{new} word embeddings, but also learning the full embedding layer (thus allowing words seen during offline training to shift their meaning).
Additionally, we test what happens when the full encoder weights are unfrozen, allowing to adapt not only the embeddings but also how they are processed sequentially.
In the latter two cases, we incorporate $L_2$ regularization over the embeddings and the model weights.

Human users interact with the system by asking
it in their own language to perform transformations on the colored
block piles, providing immediate feedback on what was the
intended target configuration.\footnote{In our experiments, we use pre-recorded data
from \cite{Wang:etal:2016}.}
In our system, each new example that the model observes is added to a buffer $B$. Then, the model is further trained with a fixed number of gradient descent steps $S$ on predicting the correct output using examples randomly drawn from a subset $B_\text{TR} \subseteq B$ of this buffer.

In order to reduce the impact of local minima that the model could encounter when learning from just a handful of examples, we train $k$ different copies (rather than training a single model) each with
a set of differently initialized embeddings for new words. 
In this way, we can pick the best model to make a future prediction, not only
based on how well it has fitted previously seen data, but also by how well it
generalizes to other examples.
For choosing which model to use, we use a different (not necessarily disjoint)
subset of examples called $B_\text{VA} \subseteq B$.
We experimented with two model selection strategies: \textbf{greedy}, by which we pick the 
model with the lowest loss computed over the full training buffer examples
($B_\text{VA} = B_\text{TR} = B$); and \textbf{1-out}, where we save the last example 
for validation and pick the model that has the lowest loss on that example 
($B_\text{VA} = B[\texttt{LAST}]$, $B_\text{TR} = B[0:\texttt{LAST}-1]$)
\footnote{Other than these, there is wealth of methods in the literature for
model selection (see, e.g. \citeauthor{Claeskens:etal:2008}, \citeyear{Claeskens:etal:2008}). To limit the scope of this work, we
leave this exploration for future work.}. Algorithm \ref{alg:online} summarizes
our approach.

\begin{algorithm}
    \small
\caption{Online Training}\label{alg:online}
\begin{algorithmic}[1]

  \State Initialize models $m_1,\dots,m_k$
  \State Let $B$ be an empty training buffer
  \For{\textit{t} = 1,2,...,\textit{T}}
    \State Observe the input utterance $\mathbf{w}_t$ and block configuration $\mathbf{x}_t$
    \State \Call{Select}{} best model $m_i$ using data $B_\text{VA}$ 
    \State Predict $\hat{\mathbf{y}_t} = m_i(\mathbf{w}_t, \mathbf{x}_t)$
    \State Observe feedback $\mathbf{y}_t$.
    \State Record prediction accuracy ($\mathbf{y}_t == \hat{\mathbf{y}}_t$)
    \State Add $(\mathbf{w}_t, \mathbf{x}_t, \mathbf{y}_t)$ to $B$
    \State \Call{Train}{} $m_1,\dots,m_k$ on data $B_\text{TR}$
  \EndFor
    \Procedure{Select}{$m_1,\dots,m_k$, $B_\text{VA}$}
    
            \State Let $C_i \gets \sum_{\mathbf{(\mathbf{w},\mathbf{x},\mathbf{y})} \in B_\text{VA}}{
            \mathcal{L}(\mathbf{y}, m_i(\mathbf{w},\mathbf{x}))}$
            \State \Return $m_i$ having minimal $C_i$
    \EndProcedure
    \Procedure{Train}{$m_1,\dots,m_k$, $B_\text{TR}$}
            \For{$i=1,\dots,k$, $s=1,\dots,S$}
                \State Draw $\mathbf{w},\mathbf{x},\mathbf{y} \sim B_\text{TR}$
                \State Predict $\hat{\mathbf{y}_t} = m_i(\mathbf{w}_t, \mathbf{x}_t)$
                \State Compute $\nabla_{\theta_i}\mathcal{L}(\mathbf{y}, \hat{\mathbf{y}_t})$
                \State Update $m_i$ by a gradient step on $\theta_i$
            \EndFor

    \EndProcedure
\end{algorithmic}
\end{algorithm}

\section{Experiments}
We seek to establish whether we can train a neural network system to learn the rules and 
structure of a task while communicating with a scripted teacher 
and then having it adapt to the particular nuances of each human user.
We tackled this question incrementally. First,
we explored what is the best architectural choice for solving the SHRDLURN task
on our large artificially-constructed dataset.
Next, we ran multiple controlled experiments to investigate the adaptation
skills of our online learning system. In particular, we first tested whether the
model was able to recover the original meaning of a word that had been replaced
with a new arbitrary symbol -- e.g. ``\emph{red}'' becomes ``\emph{roze}''--
on an online training regime. Finally, 
we proceeded to learn from real human utterances using the dataset collected by
\cite{Wang:etal:2016}. 
 
\subsection{Offline training} \label{sec:pretraining}
We used the data generation method described in the previous section
to construct a dataset to train our neural network systems. To evaluate
the models in a challenging compositional setting, rather than producing a random split of the data, we create validation and test sets that have no overlap with training instructions or block configurations.
To this end, we split all the 88 possible utterances that can
be generated from our grammar into 66 utterances for training, 11 for validation and 11 for testing. 
Similarly, we split all possible 85 combinations that make a valid
column of blocks into 69 combinations for training, 8 for validation and 8 for testing,
sampling input block configurations using combinations of 6 columns pertaining
only to the relevant set. In this way, we generated 42000 instances for training, 4000 for validation and 4000 for testing.

We explored all possible combinations of encoder and decoder models%
: LSTM encoder and LSTM decoder
(\textbf{seq2seq}), LSTM encoder and convolutional decoder (\textbf{seq2conv}),
convolutional encoder and LSTM decoder (\textbf{conv2seq}), and both
convolutional encoder and decoder (\textbf{conv2conv}). Furthermore, we explored
a bag of words encoder with an LSTM decoder (\textbf{bow2seq}).  We trained 5 models with our generated
dataset and use the best
performing for the following experiments. 
We conducted a hyperparameter search for all these models, exploring the number
of layers (1 or 2 for LSTMs, 4 or 5 for the convolutional network), the size of 
the hidden layer (32, 64, 128, 256) and dropout rate (0, 0.2, 0.5). For each
model, we picked the hyperparameters that maximized accuracy on our validation
set and report validation and test accuracy in Table~\ref{tab:offline-accuracy}.

As it can be noticed, seq2conv is the best model for this task by a large margin, performing perfectly or almost perfectly on this challenging test split featuring only unseen utterances and block configurations. Furthermore, this validates our hypothesis that the convolutional decoder is better fitted to process the structure of the block piles. 

\begin{table}
    \small
    \centering
    \begin{tabular}{|c|cc|}
        \hline
        \textbf{Model} & \textbf{Val. Accuracy} & \textbf{Test Accuracy} \\\hline
        seq2seq & $78$ & $79$ \\
        seq2conv & $\mathbf{99}$ & $\mathbf{100}$ \\
        conv2seq & $73$ & $67$ \\
        conv2conv & $64$ & $74$ \\
        bow2seq & $57$ & $63$ \\ \hline
    \end{tabular}
    \caption{Model's accuracies (in percentages) evaluated on block configurations and utterances that were completely unseen during offline training. Results expressed in percentages.}
    \label{tab:offline-accuracy}
\end{table}

\subsection{Recovering corrupted words}
\label{sec:word-relearning}
Next, we ask whether our system could adapt quickly to controlled variations in the language.
To test this, we presented the model with a simulated user producing utterances drawn from the same grammar as
before, but where some words have been systematically corrupted so the model cannot recognize
them anymore. We then evaluated the model on whether it can recover the meaning
of these words during online training. 
For this experiment, we combined the validation and test sections of our dataset, containing in all 22 different utterances, to make sure that the presented utterances were completely unseen during training time. 
We then split the vocabulary in two disjoint sets of words that we want to corrupt, one for validation and one for testing. 
For validation, we take one verb (``add''), 2 colors (``orange'' and ``red''), and 4 positions (``1st'', ``3rd; ; ``5th'' and ``even''), keeping the remaining alternatives
for testing.
We then extracted a set of 15 utterances containing these
words and corrupted each occurrence of them by replacing them with a new token
(consistently keeping the same new token for each occurrence of the word).
In this way, we obtained a validation set where we can calibrate hyper-parameters
for all the test conditions that we describe below.
We further extracted, for each of these utterances, 3 block configurations to
pair them with, resulting in a simulated session with 45 instruction examples.
For testing, we created controlled sessions where we corrupted: 
one single word, two words of different type (e.g. verb and color), three words of different types and finally, all words from the test set
vocabulary\footnote{We also experimented
with different types of corrupted words (verbs, colors or position numerals) but we found no obvious differences between them.}\footnote{The dataset is available with the supplementary materials at \url{https://github.com/rezkaaufar/fast-and-flexible}.}.
Each condition allows for different a number of sessions because of the number of ways one can chose words from these sets.
By keeping the two vocabularies disjoint we make sure that by optimizing the hyperparameters of our online training
scheme, we don't happen to be good at recovering words from a particular subset.

We use the validation set to calibrate the hyperparameters of the online training routine.
In particular, we vary
the optimization algorithm to use (Adam or SGD), the number of training steps (100,
200 or 500), the regularization weight ($0$, $10^{-2}$, $10^{-3}$, $10^{-4}$),
the learning rate ($10^{-1}$, $10^{-2}$, $10^{-3}$), and the model selection
strategy (greedy or 1-out), while keeping the number of model parameters that
are trained in parallel fixed to $k = 7$. 
For this particular experiment, we considered learning only
the embeddings for the new words, leaving all the remaining weights frozen
(model \textbf{1}). To assess the relative merits of this model, we compared
it with ablated versions where the encoder has been randomly initialized but
the decoder is kept fixed (model \textbf{2}) and a fully randomly initialized
model (model \textbf{3}).
Furthermore, we evaluated the impact of having multiple ($k=7$) concurrently trained model parameters by comparing it with just having a single set of parameters trained (model \textbf{4}). 
We report the best hyperparameters for each model in the supplementary materials.  
We use online accuracy as our figure of merit, computed as
$\frac{1}{T}\sum_{t=1}^T{\mathbb{I}[\mathbf{\hat{y}}_t == \mathbf{y}_t]}$, where $T$ is the length of the session.
We report the results in Table~\ref{tab:recovering-words}.

\begin{table}[htb]
    \setlength{\tabcolsep}{5pt}
    \begin{small}
    \begin{center}
        \centering
        \begin{tabular}{lcc|cccc}
            &&&\multicolumn{4}{c}{\bf N. of corrupted words}\\\hline
            &\textbf{Transfer} & \textbf{Adapt} & \textbf{1} & \textbf{2} & \textbf{3} & \textbf{all} \\\hline
            \multicolumn{3}{c|}{$k=7$}&\multicolumn{4}{c}{}\\
            \cline{1-3}
            1.&Enc+Dec&Emb.& $90.9$ & $\mathbf{88.1}$&  $\mathbf{86.1}$ & $\mathbf{73.3}$ \\
            2.&Dec.&Enc. & $\mathbf{93.8}$ & $85.9$ & $82.4$ & $55.5$ \\
            3.&$\emptyset$ & Enc + Dec & $43.3$ & $35.5$ & $36.1$ & $36.7$ \\
            \cline{1-3}
            \multicolumn{3}{c|}{$k=1$}&\multicolumn{4}{c}{}\\
            \cline{1-3}
            4.&Enc+Dec&Emb.& $86.1$ & $84.3$&  $81.8$ & $55$ \\
        \end{tabular}
    \end{center}
    \end{small}
    \caption{Online accuracies (in percentages) for the word recovery task averaged over 7 
    sessions for 1 word, 17 for 2 words, 10 for 3 words and a single
    interaction for the all words condition, having 45 instructions each. ``Transfer'' stands for the components whose weights were saved (and not reinitialized) from the offline training. ``Adapt'' stands for the components whose weights get updated during the online training.}
    \label{tab:recovering-words}
\end{table}

First, we can see that --perhaps not too surprisingly-- the model that adapts only the word embeddings performs best overall.
Notably, it can reach 73\% accuracy even when all words have been corrupted (whereas, for example, the model of \cite{Wang:etal:2016} obtains 55\% on the same task).
The only exception comes in the single corrupted word condition, where re-learning the full encoder seems to be performing even better. A possible explanation is given by the discrepancy between this condition and the validation set, which was more akin to the ``all'' condition, resulting in suboptimal hyperparameters for the condition with a single word changed.  Nevertheless, it is encouraging to see that the model can quickly learn to perform the instructions even in the most challenging setting where all words have been changed.
In addition, we can observe the usefulness of having multiple sets of parameters trained, by comparing the ``Embeddings'' models  by default trained with $k=7$ models and when $k=1$, noting that the former is consistently better.

\subsection{Adapting to human speakers}
\label{sec:human-data}

Having established our model's ability to recover the meaning of masked known 
concepts, albeit in similar contexts as those seen seen during training, we
moved to the more challenging setting where the model needs to adapt to real
human speakers. 
In this case, the language can significantly depart from the one seen during
the offline learning phase, both in surface form and in their underlying
semantics.  For these experiments we used the dataset made available by
\cite{Wang:etal:2016}, collected from turkers playing SHRDLURN in
collaboration with their log-linear/symbolic model. The dataset contains 100
sessions with nearly 8k instruction examples. 
We first selected three sessions in this dataset to produce a validation set 
to tune the online learning hyperparameters. 
All the remaining 97 sessions were left for testing.
In order to assess the relative importance of our pre-training procedure on each
of our model's components, we explored 6 different variants of our model.
On one hand, we varied which set of pre-trained weights were kept without
reinitializing them: \textbf{(a)} All the weights in the encoder plus all the
weights of the decoder; \textbf{(b)} only the decoder weights while randomly
initializing the encoder; or \textbf{(c)} no weights and thus, resetting them
all (this taking the role of a baseline for our method). 
On the other hand, we explored which subset of weights we adapt, leaving all
the rest frozen: \textbf{(1)} Only the word embeddings\footnote{Here we report
adapting the full embedding layer, which for this particular experiment performed better than just adapting the embeddings for new words.}, \textbf{(2)} the full weights of the encoder or
\textbf{(3)} the full network (both encoder and decoder). 
Among the 9 possible combinations, we restricted to the 6 that wouldn't result
on random components not being updated (for example (c-2) would result in a
model with a randomly initialized decoder that is never trained), thus leaving
out (c-1), (c-2) and (b-1).
For each of the remaining 6 valid training regimes, we ran an independent
hyperparameter search choosing from the same pool of candidate values as in
the word recovery task (see Section \ref{sec:word-relearning}). We picked the
hyperparameter configuration that maximized the average online accuracy on the
three validation sessions. The best hyperparameters are reported on the
supplementary materials. 

\begin{table*}[ht]
    \centering
        \begin{tabular}{|c|c|c|c|c|c|c|c|}
            \cline{3-8}
            \multicolumn{2}{c|}{}& \multicolumn{6}{c|}{\textbf{Adapt}}\\\cline{3-8}
            \multicolumn{2}{c|}{}& \multicolumn{2}{c|}{(1) Embeddings} & \multicolumn{2}{c|}{(2) Encoder} & \multicolumn{2}{c|}{(3) Encoder+Decoder} \\\cline{3-8}
            \multicolumn{2}{c|}{}& acc.&$r$& acc.&$r$& acc.&$r$\\\cline{1-8}
            \parbox[t]{2mm}{\multirow{3}{*}{\rotatebox[origin=c]{90}{\textbf{Reuse}}}}&
            (c) Nothing (Random) & -&-&-&- & $13.5$& $0.58$ \\\cline{2-8}
            &(b) Decoder (Random Encoder) & -&- & $\mathbf{23}$ & $0.83$ & $21$ & $0.7$\\\cline{2-8}
            &(a) Encoder +  Decoder & $18.2$ &$0.74$& $22.6$&$\mathbf{0.84}$ & $21.3$&$0.72$ \\\cline{1-8}
        \end{tabular}
    \caption{For each (valid) combination of set of weights to re-use and
    weights to adapt online, we report average online accuracy on
    \cite{Wang:etal:2016} dataset and pearson-$r$ correlation between online accuracies
    obtained by our model and those reported by the author.
    Results obtained on 220 sessions, with about 39 $\pm$ 31 interactions each.} 
    \label{tab:human-results}.
\end{table*}

We then evaluated each of the model variants on the 97 interactions in our test
set using average online accuracy as figure of merit.
Furthermore, we also computed the correlation between the online accuracy 
obtained by our model on every single session and that obtained by
\citeauthor{Wang:etal:2016}'s system which was endowed with hand-coded functions. 
The higher the correlation, the more our model behaves in a similar fashion
to theirs, learning or failing to do so on the same sessions.

The results of
these experiments are displayed in Table \ref{tab:human-results}.

In the first place, we observe that models using knowledge acquired in the
offline training phase (rows a and b) perform (in terms of accuracy) better
than a randomly initialized model (c-3), confirming the effectiveness of our
offline training phase.
Second, a randomly initialized encoder with a fixed decoder (b-2) performs
slightly better than the pre-trained one (a-2). 
This result suggests that the model is better off ignoring the specifics of our
artificial grammar\footnote{Recall that the encoder is the component that reads
and interprets the user language, while the decoder processes the block
configurations conditioned on the information extracted by the encoder.} and
learning the language from scratch, even from very few examples.
Therefore, no manual effort is required to reflect the specific surface
form of a user's language when training the system offline on artificial data. 
Finally, we observe that the models that perform the best are those in column
(2) which adapt the encoder weights and freeze the decoder ones.
This is congruent to what would be expected if the decoder is implementing
task-specific knowledge because the task has remained invariant between the two
phases and thus, the components presumedly related to solving it should not need to
change.
Interestingly, variants in these column also correlate the most with the
symbolic system. 
Moreover, performance scores seem to be strongly aligned with the correlation
coefficients. As a matter of fact the 7 entries of online accuracy and
pearson $r$ are themselves correlated with $r=0.99$, which is highly
significant even for these few data points.
This result is compatible with our hypothesis that the symbolic system carries
learning biases which, the better our models are at capturing, the better they
will perform in the end task. 
Still, evidence for our hypothesis, based both on the effectiveness of the
pre-training step and on the fact that similar
systems should succeed and fail on similar situations, is still indirect. 
We leave for future work the interesting question of through which mechanisms
the decoder is implementing useful task-specific information, and whether they
mimic the functions that are implemented in \citeauthor{Wang:etal:2016}'s 
system, or whether they are of a different nature.

Furthermore, to test whether the model was harnessing similarities between our
artificial and the human-produced data, we re-trained our model on our
artificial dataset after scrambling all words and shuffling word order in all
sentences in an arbitrary but consistent way, thus destroying any existing
similarity at lexical or syntactic levels.
Then, we repeated the online training procedure keeping the decoder weights,
obtaining 20.7\% mean online accuracy, which is much closer to the results of
the models trained on the original grammar than it is to the randomly
initialized model.
With this, we conclude that a large part of the knowledge that the model
exploits comes from the tasks mechanics than from specifics of the language
used.

Finally, we note that the symbolic model attains a higher average online
accuracy of $33\%$ in this dataset, showing that there is still room for
improvement in this task. 
Yet, it is important to remark that since this model features hand-coded domain
knowledge it is expected to have an advantage over a model that has to learn
these rules from data alone, and thus the results are not directly comparable
but rather serve as a reference point.

\section{Analysis}
\begin{figure}
    \nocaption
    \begin{subfigure}[b]{0.45\textwidth}
    \centering
    \includegraphics[width=190pt]{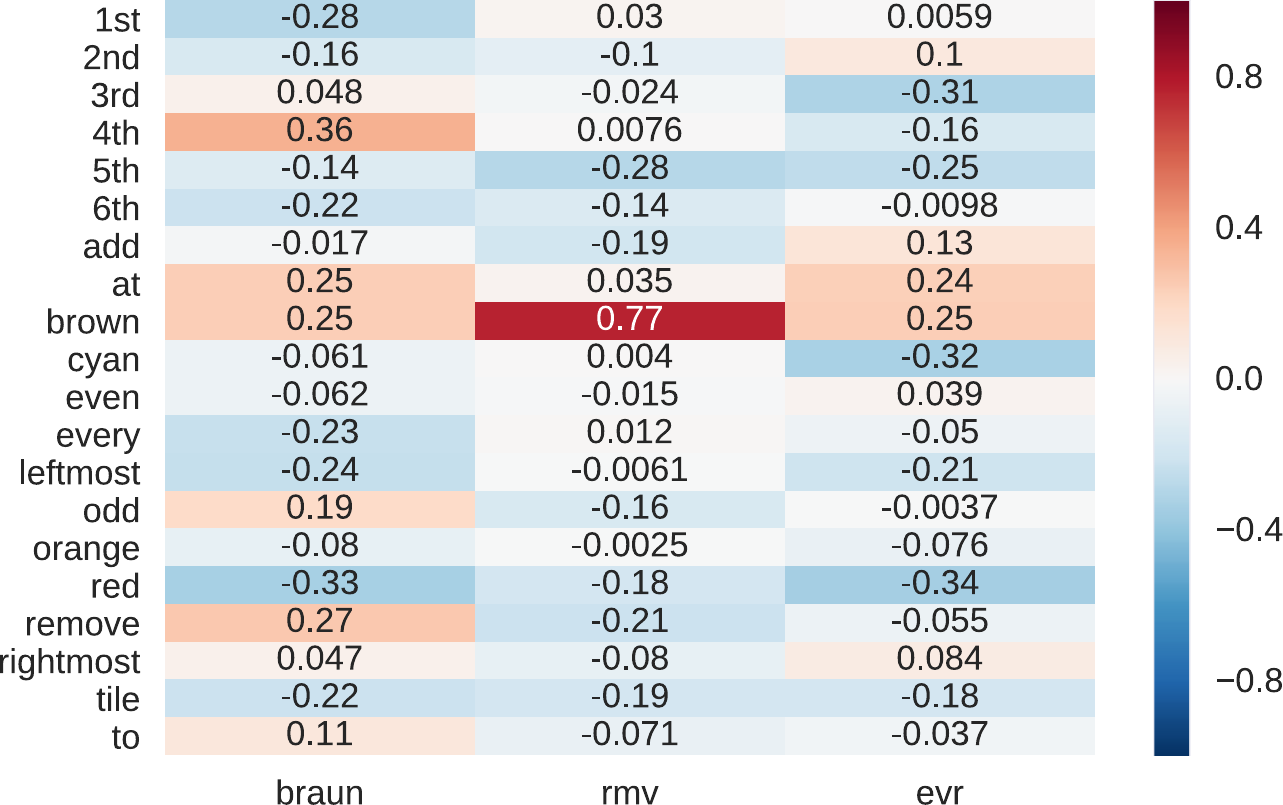}
    \caption{Cosine similarities of the newly learned word embedding for the corrupted version of the words ``brown'', ``remove'' and ``every'' with the rest of the vocabulary.}
    \label{fig:rmv-braun-evr}
    \end{subfigure}
    \hfill
    \begin{subfigure}[b]{0.45\textwidth}
    \centering
    \includegraphics[width=0.75\linewidth]{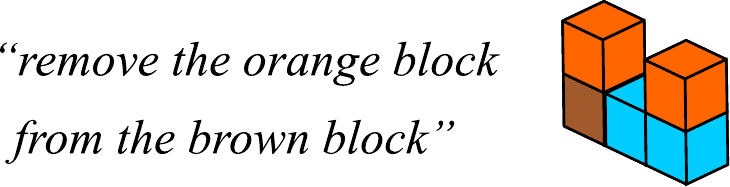}
\caption{Example of failing case for our system. During offline training it had not seen other colored blocks to be used as referring expressions for locations.}\label{fig:failing-example}
    \end{subfigure}
\end{figure}

\label{sec:analysis-recovery}

\paragraph{Word recovery}
To gain some further understanding of what our model learns, we examined the word embeddings learned by our model in the word recovery task. 
In particular, we wanted to see whether the embedding that the model had re-learned for the corrupted word was similar to the embedding of the original word.
We analyzed a session in which 3 words had been corrupted: ``brown'', ``remove'' and ``every''.
Recall from Section \ref{sec:word-relearning} that these sessions are 45-interactions-long with 15 different utterances issued on 3 different inputs each.
We then evaluated how close each of the corrupted versions of these words (called ``braun'', ``rmv'' and ``evr'') were to their original counterparts in terms of cosine similarity. 
Interestingly, the model performs very well, with an online accuracy of about 80\%, with 50\% of the errors concentrated on a single utterance that contains all corrupted words together: ``rmv braun at evr tile''.
However, as shown on Figure \ref{fig:rmv-braun-evr}, the system seems to be assigning most of the semantics associated to ``brown'' to the embedding for ``rmv'' (``brown'' has much higher cosine similarity to ``rmv'' than to ``braun''), implying that the system is confounding these two words.
This is consistent with previous findings on machine learning systems~\citep{Sturm:2014}, showing that systems can easily learn some spurious correlation that fits the training data rather than the ground-truth generative process.
Similar observations were brought forward on a linguistic context by \citet{Lake:Baroni:2017}, where the authors show that, after a system has learned to perform a series of different instructions (e.g., ``run'', ``run twice'', ``run left''), a new verb is taught to it (``jump''), but then it fails to generalize its usage to previously known contexts (``jump twice'', ``jump left'').
While our system seems to be capable of compositional \emph{processing}, as suggested by the high accuracy on our compositional split shown in Section~\ref{sec:pretraining}, it is not able to harness this structure during learning from few examples, as evidenced by this analysis. In other words, it is not capable of compositional \emph{learning}.
One possible route to alleviate this problem could include separating syntax and semantics as is customary on formal semantic methods~\citep{Partee:etal:1990} and, as recently suggested in the context of latent tree learning~\citep{Havrylov:etal:2019}, so that syntax can guide semantics both in processing and learning.

\paragraph{Human data} On the previous section we have shown that the performance of our system correlates strongly with the symbolic system of Wang et al. 
Yet, this correlation is not perfect, and thus, there are sessions in which our system performs comparatively better or worse on a normalized scale. 
We looked for examples of such sessions in the dataset. Figure \ref{fig:failing-example} shows a particular case that our system fails to learn. 
Notably it is using other blocks as referring expressions to indicate positions, a mechanism that the model had not seen during offline training, and thus it struggled to quickly assign a meaning to it. 

On more realistic settings, language learning does not take the form of our idealized two-phase learning process, but it is an ongoing learning cycle where new communicative strategies can be proposed or discovered on the fly, as this example of using colors as referring expressions teaches us. Tackling this learning process requires advances that are well out of the scope of this work \footnote{An easy fix would have been adding instances of this mechanism to our dataset, possibly improving our final performance. Yet, this bypasses the core issue that we attempt to illustrate here. Namely, that humans can creatively come up with a potentially infinite number of strategies to communicate and our systems should be able to cope with that.}. However, we see these challenges as exciting problems to pursue in the future.

\section{Conclusions}

Learning to follow human instructions is a challenging task because humans
typically (and rightfully so) provide very few examples to learn from.  For
learning from this data to be possible, it is necessary to make use of some
inductive bias. Whereas work in the past has relied on hand-coded components or
manually engineered features, here we sought to establish whether this
knowledge can be acquired automatically by a neural network system through a
two phase training procedure: A (slow) offline learning stage where the network
learns about the general structure of the task and a (fast) online adaptation
phase where the network needs to learn the language of a new specific speaker.
Controlled experiments demonstrate that when the network is exposed to a
language which is very similar to the one it has been trained on except for
some new synonymous words, the model adapts very efficiently to the new vocabulary,
albeit making non-compositional inferences.
Moreover, even for human speakers whose language usage can considerably depart 
from our artificial language, our network can still make use of the inductive
bias that has been automatically learned from the data to learn more efficiently.
Interestingly, using a randomly initialized encoder on this task performs
equally well or better than the pre-trained encoder, hinting that the knowledge
that the network learns to re-use is more specific to the task rather than
discovering language universals. This is not too surprising given the
minimalism of our grammar.

To the best of our knowledge we are the first to present a neural model to play
the SHRDLURN task without any hand-coded components.
We believe that an interesting direction to explore in the future is adopting
meta-learning techniques \citep{Finn:etal:2017,Ravi:etal:2016}, to tune the 
network parameters having in mind that they should serve for
adaptation, or adopting syntax-aware models, which may improve sample 
efficiency for learning instructions.
We hope that bringing together these techniques with the presented here, we can
move closer to having fast and flexible human assistants.

\section*{Acknowledgments} We would like to thank Marco Baroni, Willem Zuidema,
Efstratios Gavves, the i-machine-think group and all the anonymous reviewers for their useful comments. 
We also gratefully acknowledge the support of Facebook AI Research to make this
collaboration possible. 

\bibliography{bibliography}
\bibliographystyle{chicago}
\end{document}